# Corn Yield Prediction Model with Deep Neural Networks for a Smallholder Farmer Decision Support System

**Chollette C. Olisah[1], Lyndon Smith[2], Melvyn Smith[3], Morolake O. Lawrence[4], Osita Ojukwu[5]**

[1] PrecXIMed International

[2,3] School of Engineering, University of the West of England.

[4] Department of Computer Science, Baze University, Abuja, Nigeria.

[5] Morph Innovations Limited, Lagos, Abuja, Nigeria.

**Abstract**

Crop yield prediction has been modeled on the assumption that there is no interaction between weather and soil variables. However, this paper argues that an interaction exists, and it can be finely modelled using the Kendall Correlation coefficient. Given the nonlinearity of the interaction between weather and soil variables, a deep neural network regressor (DNNR) is carefully designed with consideration to the depth, number of neurons of the hidden layers, and the hyperparameters with their optimizations. Additionally, a new metric, the average of absolute root squared error (ARSE) is proposed to combine the strengths of root mean square error (RMSE) and mean absolute error (MAE). With the ARSE metric, the proposed DNNR(s), optimised random forest regressor (RFR) and the extreme gradient boosting regressor (XGBR) achieved impressively small yield errors, 0.0172 t/ha, and 0.0243 t/ha, 0.0001 t/ha, and 0.001 t/ha, respectively. However, the DNNR(s), with changes to the explanatory variables to ensure generalizability to unforeseen data, DNNR(s) performed best. Further analysis reveals that a strong interaction does exist between weather and soil variables. Precisely, yield is observed to increase when precipitation is reduced and silt increased, and vice-versa. However, the degree of decrease or increase is not quantified in this paper. Contrary to existing yield models targeted towards agricultural policies and global food security, the goal of the proposed corn yield model is to empower the smallholder farmer to farm smartly and intelligently, thus the prediction model is integrated into a mobile application that includes education, and a farmer-to-market access module.

## 1    Introduction

The role of the smallholder farmer is critical to global food security. In Sub-Saharan Africa, the smallholder farmer constitutes about 80 % of crop farmers [1], yet the region is challenged by an acute food crisis. As highlighted in the 2022 Global Report on Food Crises (GRFC) mid-year update, around 140 million people in Sub-Saharan Africa are experiencing severe food insecurity. Several factors can impact smallholder food production capacities, which are summarized in this paper as secondary and primary factors, based on the level of effect on the food production capacities of the smallholder farmer. The secondary-level factors result from the impact of agricultural and rural policies [2] and weather variability [3] on food production capacities. This level cannot be directly controlled by the smallholder farmer. The primary-level factors, on the other hand, can be directly controlled by the farmer; examples are lack of access to the market, poor education, and lack of technology for precision farming [4]. This paper argues that when technology is leveraged, the smallholder farmer can overcome the challenges of the primary-level factors, which could potentially increase food production and impact food security in Africa.

Technology in the form of robots [5], sensors [6], drones [7], and decision systems [8] – [16] are currently enabling radical transformations in precision agriculture. Aside from a decision system that can be designed with inexpensive predictive models, other technologies are costly and might be out of the reach of the smallholder farmer. As a result, subsequent discussions will review ways decision systems have been explored in literature for smart farming.

The decision systems are approached in the existing literature from the perspective of crop yield prediction which is aimed at helping governments to monitor food production, improve agriculture policies, and monitor food security. Jeong *et al.* [11] trained the random forest (RF) machine learning algorithm for crop yield prediction. Alhnaity *et al.* [12] applied long short-term memory (LSTM) to predict tomato yield in a controlled environment. Using 142,952 samples of maize data comprising plant genotypes, weather, and soil variables, [13] designed a deep neural network (DNN) algorithm for yield prediction. Their DNN architecture comprised 21 hidden layers with 50 neurons in each layer. In [14] two convolutional neural networks (CNN), which they termed weather CNN (W-CNN) and soil CNN (S-CNN), are designed for modelling temporal and spatial information on weather and soil data. The spatial information is retrieved from the fully connected (FC) layer of W-CNN and S-CNN. The spatial data, along with the temporal data, historic yield, and management data, are fed to a recurrent neural network (RNN) for forecasting yield. Similarly, Shahhosseini *et al.* [15] used a W-CNN and S-CNN to model weather and soil, and a DNN for management data. These models are used to construct a homogeneous ensemble model for corn yield prediction. An ensemble machine learning model is created in [16] by combining the following, linear regression, least absolute shrinkage, selection operator (LASSO) regression, extreme gradient boosting (XGBoost), light gradient boosted machine (lightGBM), and RF, for predicting corn yield.

Aside from the fact that existing work approaches crop yield prediction in a way that benefits the government and commercial farms more than the smallholder farmer, some of the work in the design of yield predictive models [14]-[16], appears to have assumed that the environmental variables such as soil and weather are independent of each other. However, the interaction between climate and soil is an integral part of plant growth [17], [18]. Additionally, there is no existing tool that overcomes the primary-level challenges such as farmer education, access to the market, and predictive models, that directly impact the smallholder farmer. Therefore, this paper proposes a decision support system as shown in Fig. 1 for the smallholder farmer with contributions as follows.

- Devise a preprocessing pipeline to address the challenges of the collected real-world weather and soil data. The challenges are inconsistencies in year intervals of the environment variables, missing data, inaccessibility of some environment variables to smallholder farmers, and yield and cultivation data with outcomes useful for commercial farming.

- Propose a novel approach to deep neural network regressor (DNNR) architecture design that better learns the dynamic interactions between soil, weather, and farm geographical location for crop yield prediction. The model is designed through careful consideration of depth, number of neurons of the hidden layers, and hyperparameters using optimization methods. This is to show how the structure of the DNN architecture can be leveraged to achieve performance comparable to tree-based models and tabular data regression tasks.

- Optimize decision-tree-based models, RFR and XGBR, through hyperparameter optimization using the grid search method. These models are known for their impressive performance on tabular data and, thus are presented as good baselines for comparisons to the deep learning-based models.

- Devise a new regression metric termed average of absolute squared error (ARSE) that combines the strengths of RMSE and MAE. Using these metrics, the investigation into the sensitives of the decision-tree-based models and the DNN-based models will be evaluated for their bias towards high cardinal features the state variables may introduce to the dataset in this paper's yield regression task.

- Additionally, a new generalization evaluation using unforeseen samples is presented to evaluate the impact of unexpected changes in sampled data points resulting from the impact of climate change on weather and soil compositions. This test can be further used to test for the susceptibility of the models to high cardinality features.

- Development and implementation of a mobile application, integrating the proposed predictive model(s) alongside farmer education module and market access module for a comprehensive decision system.

The proposed decision support system provides smallholder farmers with inexpensive ways to farm smart via a smartphone that is readily available to farmers while utilizing locally sourced information on the farm to make informed decisions for their farmland during a planting season. This paper only focuses on single-point yield for the corn plant. The remaining parts of the paper are structured as follows: Section 2 describes the research methodology, and Section 3 presents the experimental settings, results, and discussions. Then finally, Section 4, is the conclusion which summarizes the findings of the paper.

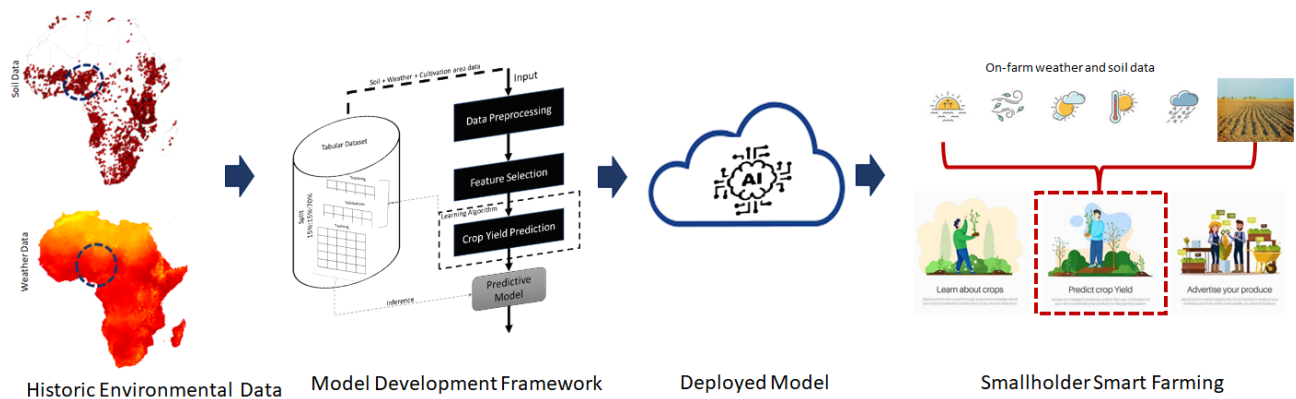

Figure 1: The pipeline for the decision support system.

## 2    Methodology

The proposed methods are presented and discussed under the following headings: data preprocessing, feature selection, and crop yield prediction. The pipeline is illustrated in Figure 2. The process begins with the collection of raw climate, weather, and environment data for Africa with Nigeria as the reference region. Then proceeds to preprocessing the data to prepare it for corn yield prediction useful to smallholder farmers. Then, a statistical correlation technique is applied to understand the interaction between weather and soil to identify the explanatory variables contributing most significantly to the outcome variable, corn yield. This process results in the selected variables with sample point values that make up the dataset.

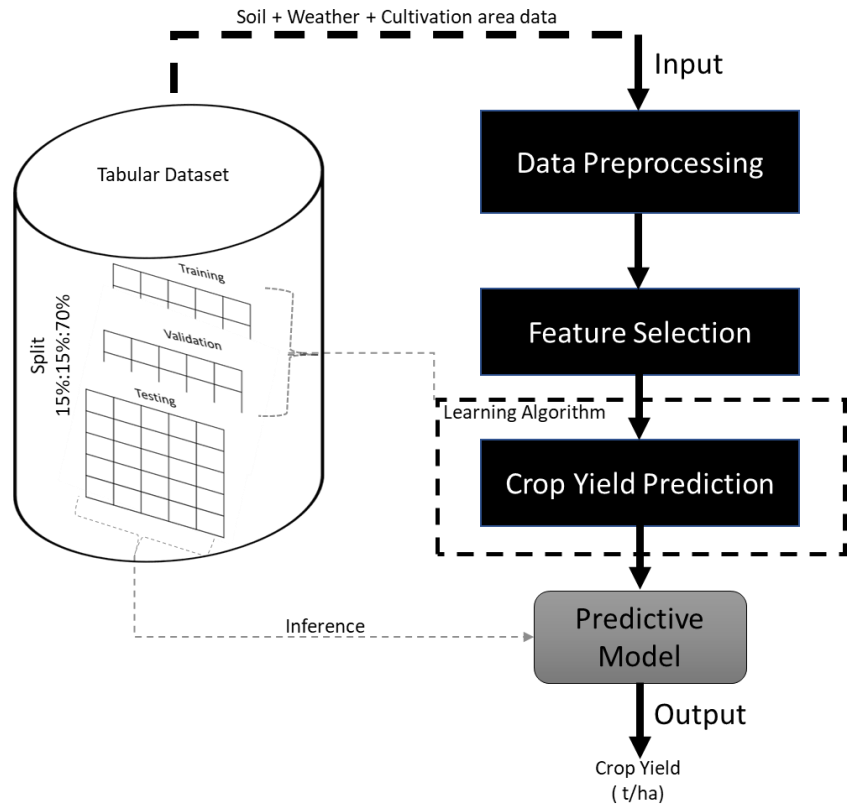

**Figure 2** The proposed corn yield prediction architecture

## 2.1    Study Region and Data Collection

As a reference study region in Africa, Nigeria is selected because there are over 211.4 million people, of which a large percentage of the population are smallholder farmers [19] who are mostly corn growers. Nigeria [9.0820° N, 8.6753° E] has an arable land area of 34 million hectares [19] and is located on the west coast of Africa. It comprises of 36 states with the most and least number of districts being 214 and 10, respectively. For each state, the environmental data are collected as follows.

i.    Grid map climate data – This data spans spatial resolutions between ~1 km$^2$ to ~340 km$^2$ from the high spatial resolution WorldClim global climate database [20]. Each grid point on the map is monthly data from January to December between 1970 and 2000 years and records 8 climate variables. The variables are average temperature $C^0$, minimum temperature $C^0$, maximum temperature $C^0$, precipitation ($mm$), solar radiation ($kJ\ m^{-2}\ day^{-1}$), wind speed ($m\ s^{-1}$), and water vapor ($kPa$) taken at 30 seconds (s), 2.5 minutes m, 5 m, and 10 m.

ii.    Grid map soil data – This data is obtained from 250 minutes of spatial resolution AfSIS soil data [22] from year 1960 to 2012. The variables are, wet soil bulk density, dry bulk density (kg dm$^{-3}$), clay percentage of plant available water content, hydraulic conductivity, the upper limit of plant available water content, the lower limit of, organic matter percentage, pH, sand percentage (g 100 g$^{-1}$), silt percentage (g 100 g⁻¹) and, clay percentage (g 100 g⁻¹), and saturated volumetric water content variables measured at depths 0–5, 5–10, 10–15, 15–30, 30–45, 45–60, 60–80, 80–100, and 100–120 measured in centimeters (cm).

iii. Corn yield data – This data is available on Kneoma Corporation website [21]. It ranged from years 1995 to 2006 and consisted of a corn yield of *1000 metric tonnes* and a cultivation area of *1000 hectares*.

iv. Geolocation coordinates (latitude and longitude) – The geolocation of each of the 36 states with their districts is sampled from Google Maps. The output feeds into the Esri-ArcGIS 2.5, a professional geographical software, for extracting the point-cloud values of each environmental variable (weather and soil) at specific geolocation of the 36 states of Nigeria.

## 2.2    Data Preprocessing

The environmental data: weather, soil, cultivation area, and crop yield data required to be preprocessed due to the following reasons, 1) inconsistency in year intervals of the environment variables because they are acquired from different sources. Particularly, the crop yield data and soil data covered periods between 1995 to 2006, and 1960 to 2012, respectively, 2) some weather and soil variables are inaccessible to smallholder farmers, and 3) The weather, soil, and cultivation area data contained missing data for some districts of a state and in some cases, the entire state.  Therefore, the following preprocessing approaches are adopted to address the challenges identified.

i. Aggregate weather and soil data across resolutions. Using the monthly weather data across the spatial resolutions of 30 s, 2.5 m, 5 m, and 10 m, compute the average of all climate data. Likewise, aggregate the soil data across depths 0–5, 5–10, 10–15, 15–30, 30–45, 45–60, 60–80, 80–100, and 100–120 in centimeters (cm).

ii. Forecast the yearly crop yield data *tonnes per hectare* and cultivation area *hectares* in 6-time steps, that is, from 2006 to 2012, so it closely matches the soil data. This is on the assumption that weather variables might be inversely related to yield and the soil variables directly related to yield. To achieve this goal, the autoregressive integrated moving average (ARIMA) [23] is used. Then, the resulting values are averaged into a single value for a state.

iii. Merge the climate, soil, yield, and cultivation area data into a single set. This step helped to reveal the states or districts with missing values. After the missing values are removed, the number of states is reduced from 36 to 23. The states' names are, Abia, Abuja, Akwa Ibom Anambra, Bayelsa, Benue, Cross River, Delta, Ebonyi, Edo, Ekiti, Enugu, Imo, Kebbi, Kwara, Lagos, Ogun, Ondo, Osun, Oyo, Plateau, Rivers, Taraba. This is proceeded by the identification of variables easily accessible to smallholder farmers as highlighted in [24] – [26].

iv. Average the yield and hectare data across years, then transform them to values realistically achievable by a smallholder farmer. The transformation functions are mathematically expressed as given in eq. (1) and eq. (2), respectively.

v. Using Kendall correlation coefficient, identify the environment (weather and soil) variables that show significant interaction with yield.

$$Y_i' = \left(\frac{Y_i}{O_t}\right) * \mathrm{E}_y \quad i = 0, 1, 2, \cdots, \mathrm{N} \tag{1}$$

$$H_i' = \left(\frac{H_i}{O_h}\right) * \mathrm{E}_h \tag{2}$$

where $Y_i$ and $H_i$ are the yield and hectare values per state, $\mathrm{E}_h$ and $\mathrm{E}_y$ the maximum expected yield and hectare values for the land capacity of a smallholder farmer. $O_h$ and $O_t$ are the original yield and hectares, respectively, in per-state capacity.

### 2.2.1 Missing Time-Step Forecasting

ARIMA [23] is a simple statistical technique for solving non-seasonal and patterned time series prediction problems. It combines auto-regressive (AR) and moving average (MA) models for forecasting future timesteps using historic observations and random errors. ARIMA is characterized by three terms: AR order term, $p$, which signifies the number of prior values to be used as predictors. The AR series stationary parameter $d$, and MA order term, $q$, indicate the number of forecast errors of past values. These parameters are required by the ARIMA Model to forecast future points in the series. The AR and MA models are then combined to form an ARIMA model [23]. This process is mathematically given as follows.

$$\hat{Y}_t = \alpha + \beta_1 Y_{t-1} + \beta_2 Y_{t-2} + \cdots \beta_p Y_{t-p} \, \epsilon_t + \emptyset_1 \epsilon_{t-1} + \emptyset_2 \epsilon_{t-2} + \cdots + \emptyset_q \epsilon_{t-q} \tag{3}$$

where $\alpha$ is a constant, $\epsilon_t$ is the error term at time $t$, $\beta_1, \beta_2, \beta_3, \cdots, \beta_p$ are the AR parameters, and $\emptyset_1, \emptyset_2, \emptyset_3, \cdots, \emptyset_q$ are the MA parameters combined to form $\hat{Y}_t$, the future points in the series.

Typically, to predict future data points of a time series data using the ARIMA model, the stationarity of the series data will have to be tested using the augmented Dickey-Fuller (ADF) test [27]. The null hypothesis of the ADF test by default represents the non-stationarity of the series, but if the $p$-value of the ADF test is less than the significance level ($p$=0.05) then the series is considered stationary. If the series passes the ADF test and is considered stationary, then the values of $p$ and $q$ terms can be determined by observing $p$-values above the significance level for the autocorrelation function and partial autocorrelation plots [27], respectively.

The crop yield and cultivation area data are extended by six (6) time steps to closely match the years of the soil data. The outcome of forecasting for some states is shown in Figure 3. The outcome of the ADF test, partial autocorrelation, and autocorrelation functions, are useful for extending the data by some data points. The importance of the ADF test is demonstrated as follows. Using the yield and cultivation area for 3 geographical states of Nigeria, Ogun, Anambra, and Taraba states with their $p$, $d$, $q$ terms given as (5,0,1; 5,0,1), (2,1,1; 1,0,1), and (2,1,1;10,1), the $p$-values given as (0.000000; 0.000000), (0.343881; 0.997572), and (0.450462, 0.859081), respectively, can be used to interpret the ADF test. Based on the significance test using the $p$-values, the Ogun state series is an example of a stationary series while the Anambra and Taraba states are examples of non-stationary series. Therefore, they will require differencing to convert them to stationary series. It should be noted that the cultivation area data is normalized using logarithmic transformation before forecasting with ARIMA because it presents highly skewed data and is reversed afterward.

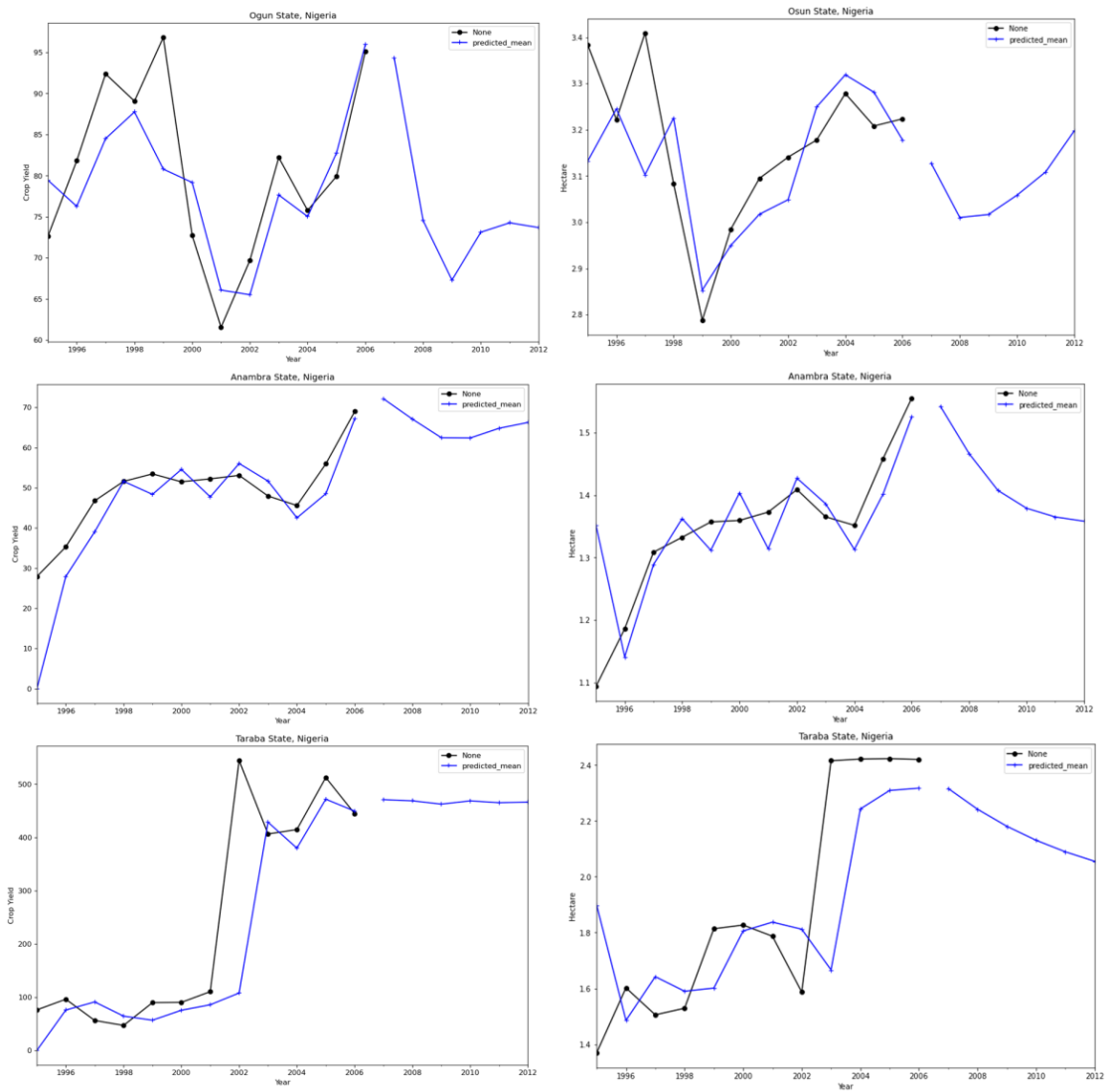

**Figure** 3 Output of forecasting future timesteps of historic yield and cultivation area using ARIMA for Anambra, Osun, and Taraba states. Figure best viewed when zoomed into

**2.3 Feature Selection**

Feature selection is a key step in machine learning (ML) for identifying intrinsic features in a dataset. This prevents ML models from overfitting to noise and can generalize well to unseen data. To select significant features, the Kendall correlation [28] is utilized. Kendall correlation is a non-parametric statistical correlation technique that measures the strength of association between two variables. This functionality makes Kendall correlation a useful tool for describing the interaction between crop yield, $y$, and any of the environment variables, $X$.

$$\tau_{x,y} = P\left(\sum_{i<j} sgn(x_i - x_j) \cdot sgn(y_i - y_j)\right) \qquad (4)$$

where $n$ is the size of the variables $(x, y)$ under observation which does not necessarily need to be ranked or ordered as is indicated in [29] and $P$ is the total number of possible pairings of $x$ with $y$ observations given as $2/n \cdot (n-1)$ , and $sgn()$ returns the sign of a real number.

The strength of the association between two variables is given by Kendall's coefficient $\tau_{x,y}$, which ranges between the values of $-1$ and $+1$, for perfect negative and positive correlation, respectively. These coefficients are coded as color maps in the correlation plot presented in Figure 4. A value $> 0$ shows a positive relationship where both the explanatory and outcome variables increase together, whereas with a value $< 0$, an increase in one will result in a decrease in the other. A value of 0 indicates that no relationship exists between both variables. The Kendall correlation amongst other statistical techniques is used because the dataset is skewed and contains relevant outliers (see Figure 5). The Kendall correlation is applied (XLSTAT statistical 2020 software) to the numeric values with the exclusion of the categorical variable, the state variables. The correlation coefficients obtained for average temperature, average minimum temperature, average maximum temperature, average precipitation, average wind speed, pH, clay, sand, silt, crop yield, and hectare, are -0.080, -0.208, -0.073, -0.475, 0.191, 0.329, -0.046, -0.560, 1.000, 1, and 0.727, respectively. The further the coefficient of an explanatory variable deviates from 0, the higher its association with the outcome variable with 0 indicating no association. The correlation coefficients of hectare (cultivation area), silt, average precipitation, sand, soil pH, wind speed, and average minimum temperature, (organized in the other of importance) were shown to be more associated with yield. The results show that silt and hectare explanatory variables are more highly correlated to yield.

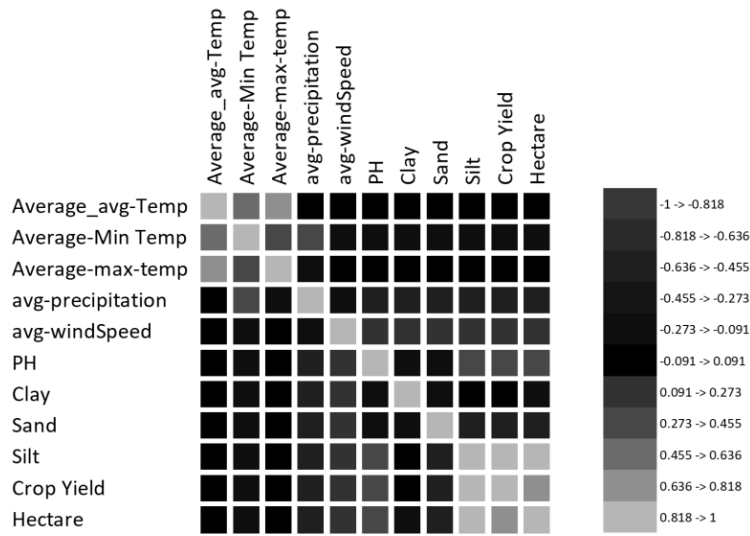

**Figure 4** Color-coded Kendall correlation coefficients. It measures the strength of association between the explanatory variables and outcome variables.

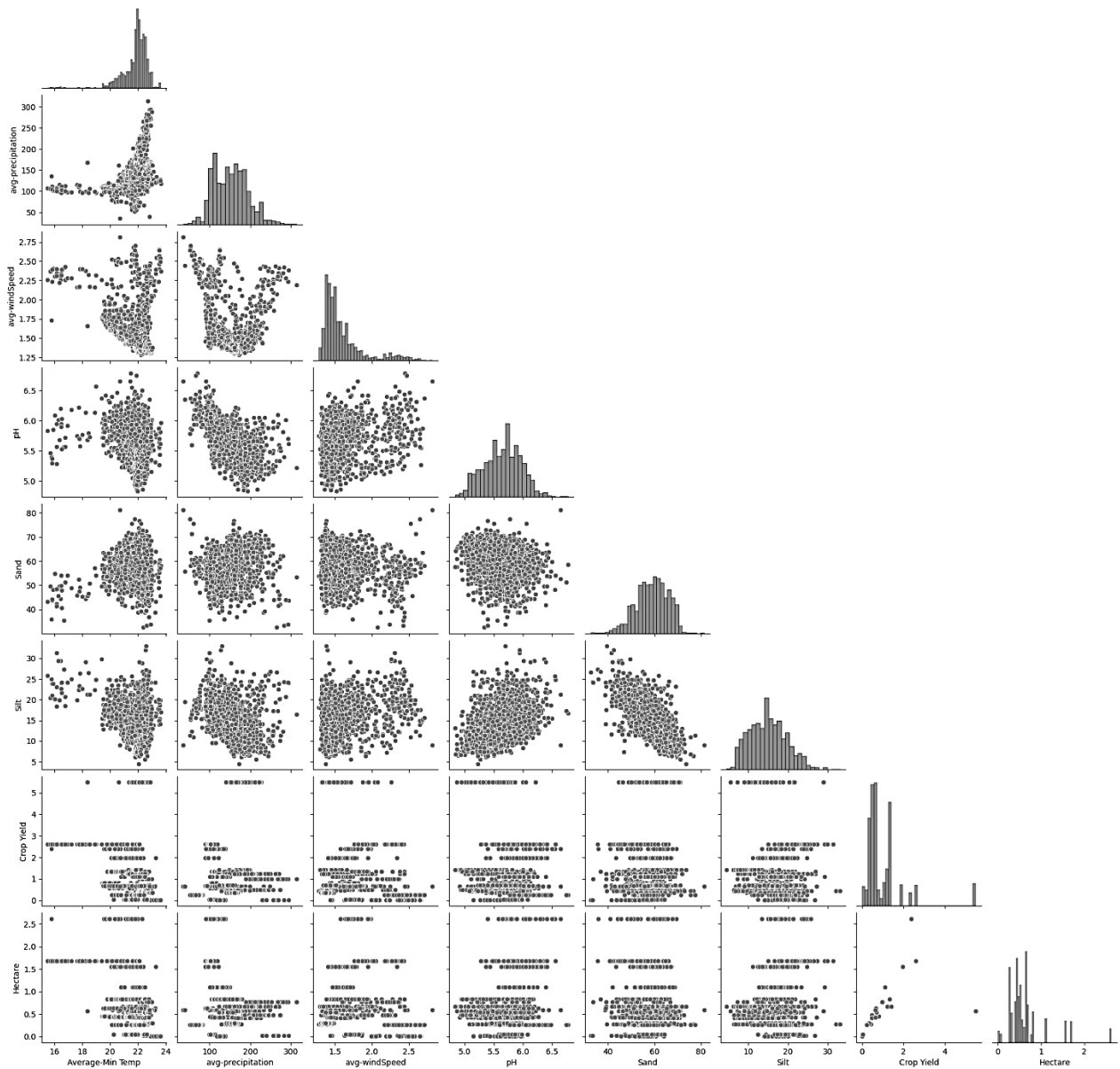

**Figure 5** Multi-variable scatter plot of the environment data distribution. Most of the explanatory variables is observed to show a nonlinear correlation to the outcome variable. Figure best viewed when zoomed into.

## 2.4 Crop Yield Prediction

The goal is to design a decision support system that maximizes efficiency by employing the predictive power of machine learning, XGBoost, and RF, which are well known to achieve high-performance accuracies on small sample tabular data explored for the regression task. In this paper, the DNN is carefully designed to significantly show usefulness in the learning of structured tabular data. The DNN architecture and hyperparameters are designed by taking into consideration the data size, its complexity, and experimental variables while XGBoost and RF are designed through hyperparameter optimization. Though these algorithms are popularly applied to classification tasks, they are utilized for this paper's regression task and will henceforth be renamed, deep neural network regressor (DNNR), extreme gradient boosting regressor (XGBR), and random forest regressor (RFR).

### 2.4.1 Deep Neural Network Regressor

The DNN is structured based on a feed-forward architecture that passes each neuron from the input layer, with associated weights and bias after transformation by nonlinear activation to the hidden layer. Then, from several hidden layers and activation functions, a decision is reached. This process is repeated using the backpropagation algorithm [30] for weight updates of the neurons until the error function is minimized. The neuron activation at the hidden layer and weight update through backpropagation are expressed as:

$$f(x) = \varphi(\sum_i^n w_i x_i + b) \tag{5}$$

where the function, $f$, for an input neuron, $x_i$, outputs a decision to pass the neuron from one hidden layer to the next. This decision is obtained through a weighted sum of neurons, $x_i$ and its weight, $w_i$, together with the added bias, $b$, and it is mapped to a desired range using the activation function, $\varphi$.

The hidden layer is the powerhouse of a DNN algorithm because the depth and number of neurons of the layer amongst other hyperparameters determine the capability of the network to address the complexity of the problem solved. The more complex the problem, the deeper the depth of the hidden layer, and vice versa. Also, with tabular data, the number of neurons can be determined from the number of variables and the size of the data. It has been shown in [31] that when the number of neurons is set to twice the number of variables in tabular data, the network begins to learn the intrinsic information of the data. The same concept is adopted in this paper; however, it is approached with a twist to how it transcends through the hidden layers. Since the complexity cannot be easily determined by the nonlinearity of the data and the number of variables, the grid search method is relied on for choosing the optimal depth of the hidden layers and other hyperparameters. In all, the proposed DNNR architecture (see Figure 6) is designed as follows: input layer - 30 neurons, hidden layer - 3 and 64 for the depth and number of neurons per layer, and output layer - a single numeric value that represents crop yield. The hyperparameters are rectified linear unit (ReLU) activation functions [32] which are applied to Eq. 5 and have been shown in the literature to be best used for overcoming the vanishing gradient problem of a network. The learning rate is set to 0.001 using the Adam optimizer which controls the weight update of the network. Epoch is set to 60 to enable the network to make 60 passes through the entire training set, in a batch size of 100, during which weight update to a neuron is made 12 times. The loss function used during weight update (backpropagation) is the mean absolute error (MAE) which is expressed as:

$$L(\emptyset) = \frac{1}{B} \sum_{i=1}^{B} |y_i - \hat{y}_i| \tag{6}$$

where $L$ is the objective function with error determined by computing the difference between $\hat{y}_i$, the predicted and $y_i$, the true value and $B$ is the total number of data points in each batch. The error value becomes input to the backpropagation computation process, given as:

$$w_x^* = w_x - \left( r * \left( \frac{\partial Error}{\partial w_x} \right) \right) \tag{7}$$

where the first term includes the weight of an input neuron $w_x$, and the last term includes the partial derivative of the error function $\left( \frac{\partial Error}{\partial w_x} \right)$ multiplied with the learning rate, $r$. By subtracting the first term from the second term, the weight for a given input neuron can be updated.

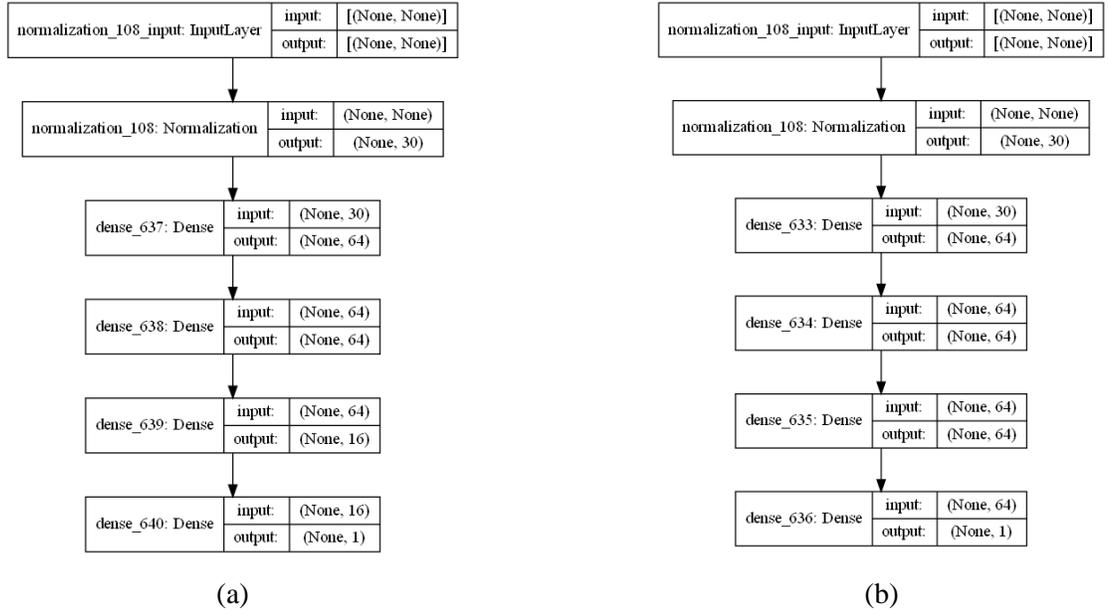

(a) (b)

**Figure 6** The architectures of the DNNs. The architecture for DNN16 (a) and DNN64 (b) both include a data normalization layer.

### 2.4.2  Extreme Gradient Boosting Regressor

The XGBoost [33] is developed based on the concept of the gradient boosting decision tree (GBDT) algorithm [34]. It builds multiple decision trees that are organized in sequences such that a preceding tree error helps minimize the prediction error of subsequent trees. In this way, a strong model of decision trees is formed. XGBoost advances this functionality by introducing the L1 and L2 regularizers to stabilize the generalization capability of GBDT. This is expressed mathematically as:

$$L(\emptyset) = \sum_{i=0}^{n} l(\hat{y}_i, y_i) + \sum_{k=0}^{n} \Omega(f_k) \tag{8}$$

where the objective of function, $L(\emptyset)$ is to impose a penalty on erroneous predictions in order to minimize the error of the model by adding a regularization function, $\Omega(f_k)$ to the loss function $l(\hat{y}_i, y_i)$ computed between the model's predicted value, $\hat{y}_i$, and actual value, $y_i$.

To predict a continuous numeric value, the negative mean average error loss function is used, thus XGBoost is referred to as XGBR. Modification to the XGBR is via hyperparameter optimization with values of the hyperparameters determined through the grid search method. For explicit discussions on the concept of the grid search method and how it is used in this paper, the reader is referred to [31]. Through grid search, the number of estimators, number of trees, maximum depth, learning rate, and minimum samples split XGBR hyperparameters are described in Table 1.

Table 1 Hyperparameters and values of the decision tree-based models.

| Hyperparameter | Description | Model Values | |
| --- | --- | --- | --- |
| | | XGBR | RFR |
| Number of estimators | Number of trees created from the training data | 900 | 10 |
| Maximum depth | Controls how specialized each tree is to the training data. The higher the value the more likely overfitting will occur. | 10 | 10 |
| Learning rate | Controls the pace at which new trees can make corrections to the error of previous trees. | 0.1 | |
| Minimum samples split | Specifies the minimum number of samples required at a leaf node for splitting to occur. | 0.1 | |
| Subsample | This signifies the number of training samples XGBR uses to grow the trees. | 1 | |

### 2.4.3 Random Forest Regressor

The RF is an ensemble of several decision trees [35] built through random sampling of training data. It uses the concept of bagging and feature randomness to ensure the decision trees each time are presented with unique samples of the training data. In this way, variance can be decreased without an increase in bias. For the regression task, the mean average error loss function is used to predict a continuous numeric value. Therefore, RF becomes a random forest regressor (RFR), and prediction from all individual trees generated from the random sample of the data, $x$, gets averaged over all trees $t_i$. The process is expressed mathematically as:

$$\hat{t} = \frac{1}{N}\sum_{i=1}^{N} t_i(x) \tag{9}$$

where *x* is the training data and *N* is the number of sets randomly created through bagging.

The RFR is optimized for the given task via the grid search method employed for the purpose of selecting the optimal values of the hyperparameters which are as follows, number of estimators, maximum depth, minimum sample split, and minimum samples leaf.

## 3    Experiments, Results, and Discussions

### 3.1    Experimental Settings

The data used in this paper comprises explanatory and outcome variables. The explanatory variables are the average minimum temperature, average precipitation, average wind speed, soil pH, soil sand content, soil silt content, and cultivation area, while the crop yield is the outcome variable. The explanatory variable is extended to include the state variable, which is a categorical data type converted to numerical through one-hot encoding. This further expands the number of explanatory variables to 30 – a combination of weather, and soil variables added to the 23 states. The inclusion of the state variable is to ensure that geolocation, which introduces changes in weather and soil composition, contributes to the decision of the predictive models.  In all, the dataset comprised 1827 data points and after missing values and duplicates were removed, only 1632 datapoints made up the dataset. The data is split in the ratio of 80:20 with 80 % for training and the remaining 20 % split in half for the validation and test sets. In all, these sets total 1142, 245, and 245 data points, respectively. For normalizing the DNNR sets, the min-max normalizer is used but all the models benefitted from cross-validation with replacement.

Given that this paper solves a regression problem, the regression metrics suffice. The RMSE is chosen so that the performance of the corn yield predictive models from the viewpoint of sensitivity to outliers (RMSE) can be observed. Usually, if there exists a data point with a large difference between the predicted and the actual, the RMSE error metric will capture it and record higher errors. However, since the RMSE is sensitive to outliers, it becomes necessary to also introduce the MAE. The MAE has its fair share of shortcomings. It is insensitive to outliers because it is unable to reveal the disparity between the actual and predicted value even when there are sample points with large errors. In this paper, both metrics are combined through averaging to utilize their benefits and overcome their shortcomings. This is on the principle that averaging is generally used by the ML community to reduce variance between several model's predictions. The new metric is termed the average of absolute root squared error (ARSE). These metrics are mathematically defined in Eq. (10-12).

To evaluate each predictive model's performance, the experiments will be presented and discussed under 1) the predictive models' overall performance with the training set, 2) the predictive model performance using different evaluation metrics on the test set, 3) the generalization capability of the predictive models on the test set, and 4) evaluation of the significance of the feature selection step in the pipeline using the test set.

$$M_{RMSE} = \sqrt{\frac{1}{N}\sum_{i=1}^{N}(y_i' - y_i)^2} \tag{10}$$

$$M_{MAE} = \frac{1}{N}\sum_{i=1}^{N}|y_i' - y_i| \tag{11}$$

$$ARSE = \frac{1}{n}\sum_{i=1}^{n}(M_{RMSE}, M_{MAE}) \tag{12}$$

where $N$ is the total number of test set data points, $y_i'$ is the predicted and $y_i$ is the true value, $M_{RMSE}$ and $M_{MAE}$ are the predicted errors for RMSE and MAE, and $n$ is the number of errors.

## 3.2 Results

### 3.2.1 Performance of the Predictive Models on the Training Set

Since the crop yield prediction in this paper is focused on the smallholder farmer rather than on agricultural policies and global food security, the corn yield prediction outputs a single-point prediction. This means that the farmer can use information such as soil pH, sand, and silt percentages, together with temperature, and precipitation to predict the yield of a given cultivation area. This data can be locally sourced from the farm to make smart decisions on the impact of weather and soil on yield. Even though corn yield prediction has no associated risk like in clinical diagnosis, making an accurate decision is as important to the farmer as any other risk-related task. For this reason, the DNNR16, DNNR64, RFR, and XGBR predictive models are designed using single-point data. The prediction loss over epoch on the training set for the DNN-based models is shown in Figure 7, and the prediction performance of the DNN-based and tree-based models achieved through 10-fold cross-validation are presented in Figure 8.

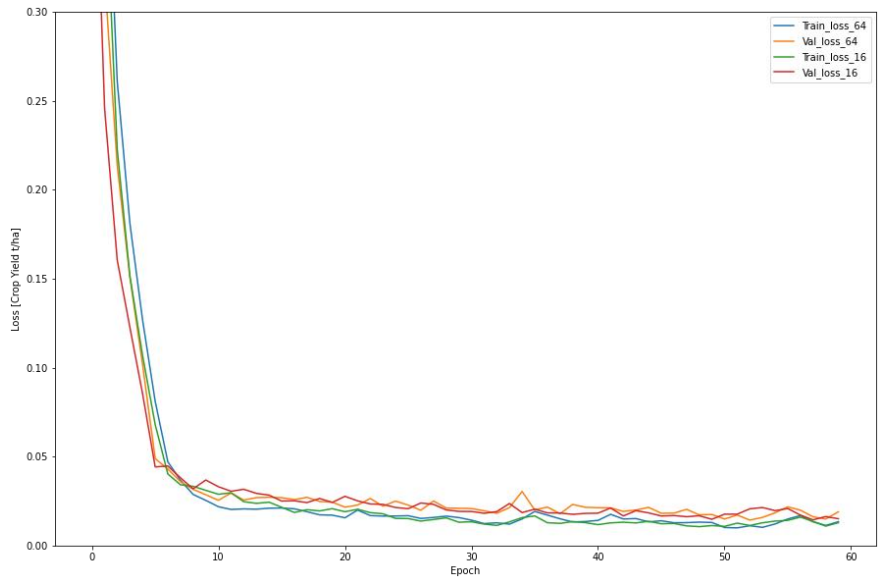

**Figure** 7 Visualizing performance on the training set. The training and validation loss of DNN16 and DNN64 are shown to be relatively the same.

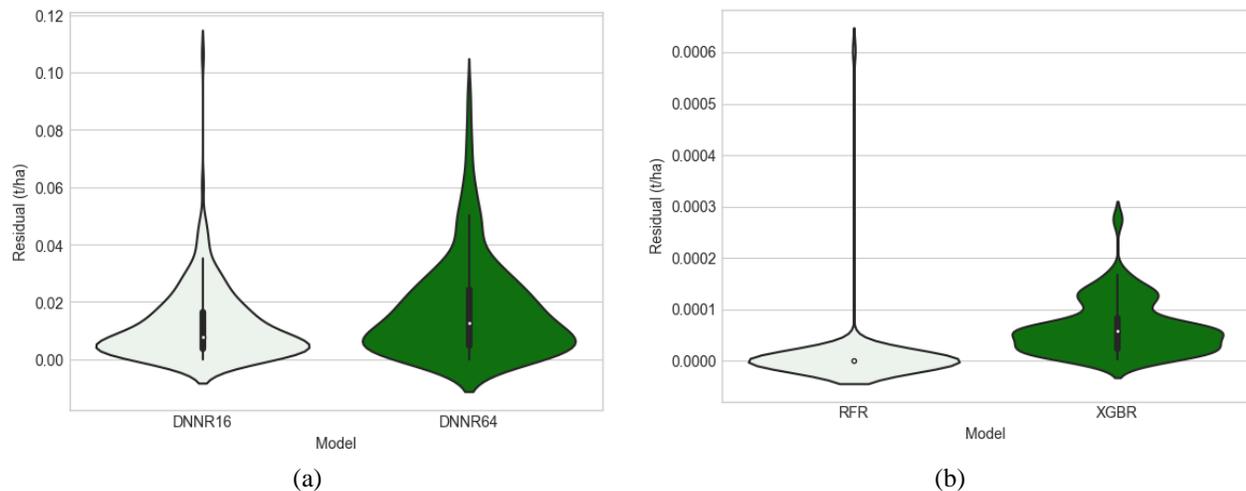

(a)                                                 (b)

**Figure 8** Visualizing distribution and probability density of training error of the predictive algorithms on the training data using violin plot. The yield errors are obtained from 10 splits of the training set with a replacement for (a) DNN-based learning algorithms and (b) decision tree-based learning algorithms.

Given that all the DNN-based models achieved very good prediction errors on the training and validation sets as can be observed in Figure 7, it can be stated that they are a good fit to the training data. Similarly, this fit is further validated through cross-validation as shown in Figure 8 using the violin plot. This is a descriptive statistical tool for visualizing a given data distribution and its probability density. Figure 8 shows that the yield errors obtained from 10 splits of the training set, with replacement, exhibited different distributions and probability densities. The DNNR64 yield errors deviated from the median of the distribution with most yield errors within 0.01 t/ha and a similar distribution can be observed for XGBR. The RFR yield errors showed *no variance*, though there is a high disparity single point outlier observed. On the other hand, the DNNR16 yield errors are skewed towards the interquartile range. Overall, the yield errors of all the predictive models are small, which demonstrates that they can learn the regression problem.

### 3.2.2 Performance on Test Set with Different Evaluation Metrics

Performance evaluation on the test set is a viewpoint of the models' performance that is critical to their generalization capability. If the models perform as well as they performed on the training set, then the models can be considered as good models for the given regression task. To evaluate the performance of the models on the test set, the RMSE, MAE, and ARSE metrics are used.

The RMSE and MAE are two popularly used metrics for evaluating the performance of regression algorithms. The outcome of using these metrics is presented in Table 2 alongside the ARSE which is expected to combine the strengths of both metrics while overcoming their shortcomings. The RMSE error is observed to be higher compared to the MAE and ARSE errors. This difference in error shows that there are sample points in the test set with large residual differences that have been flagged with RMSE. However, ARSE achieves a good balance in the errors. Interestingly, all the obtained yield errors are on average less than 0.02 t/ha which reflects how well the models are a good fit to the data. In the order of performance, RFR, XGBR, DNNR16, and DNNR64 models achieved yield prediction errors of approximately 0.0001 t/ha, 0.001 t/ha, 0.01 t/ha, and 0.02 t/ha, respectively, with ARSE metric. The RFR and XGBR models can be observed to be the best-performing models. However, their 0.001 and 0.0001 t/ha yield errors might suggest a bias towards high cardinal features. However, further analysis is needed to confirm this behavior.

Table 2 Predictive models performance across different evaluation metrics.

| Model | RMSE (t/ha) | MAE (t/ha) | ARSE (t/ha) |
|---|---|---|---|
| DNNR16 | 1.72e-02 | 1.20e-02 | 1.46e-02 |
| DNNR64 | 2.43e-02 | 1.75e-02 | 2.09e-02 |
| RFR | 5.39e-04 | 4.93e-05 | 2.94e-05 |
| XGBR | 1.43e-03 | 1.54e-04 | 7.92e-04 |

The RFR and XGBR models both originated from the family of decision tree models and, therefore are likely to share things in common, especially with respect to how the tree is generated. The decision tree-based models have a shortcoming, which is that they are sometimes biased towards high cardinal features. A cardinal feature contains a high number of unique numerical values that encode its various categorical entries. For instance, the state variable is a high cardinal feature because each of the 23 states is uniquely encoded. This means the state variable has the potential to cause a model to be heavily dependent on it for decision-making. Possibly, the cardinality problem can be higher and more likely when the explanatory variables are not strongly correlated to the outcome variable. If this is the case, then the model is likely to be strongly dependent on the high cardinality features. These types of features can hinder a model's ability to adapt to sudden and gross changes to the explanatory variables which is expected of weather and soil variables influenced by climate change.

### 3.2.3 Single-Point Generalization Assessment of the Predictive Models

To investigate how well the proposed predictive models perform on the test set, a single-point assessment of generalization is carried out using the different predictive models on the same data points from the test set. The sample data points of interest include features that describe two states, Enugu and Plateau. The Enugu state explanatory variables and values are average minimum temperature – 21.69208848 (C$^0$), average precipitation - 133.5208333 (mm), average wind speed - 1.498848967 (m s$^{-1}$), pH - 5.466666667, sand - 59.83333333 (g 100$^-$

[1]), silt - 10.1666667(g 100[-1]), and cultivation area - 0.545488917 (*hectare*) are expected to result in a yield of 0.709388681 (*tonnes per hectare*). Then the Plateau state explanatory variables and values are, average minimum temperature – 16.68546347 (C[0]), average precipitation - 99.125 (mm), average wind speed - 2.417177081 (m s[-1]), pH - 5.566666667, sand - 35.5 (g 100[-1]), silt - 27.33333333 (g 100[-1]), and cultivation area - 1.686767501 (*hectare*) are expected to result in a yield of 2.60302342 (*tonnes per hectare*). The models are observed based on 1) generalization to unseen samples, and 2) generalization of the models to unforeseen samples. The unforeseen sample is coined in this paper to represent the sudden change in values of an explanatory variable. This type of test is rarely done in literature, but it is necessary because it has strong implications for real-world use.

Figure 9 (a) illustrates the models' prediction errors observed based on (1) for Enugu and Plateau states. The DNN-based models, DNNR16 and DDNR64, are observed to have achieved residual errors of 0.0426 t/ha and 0.0429 t/ha, respectively, while the decision tree-based models, RFR and XGBR, achieved residuals of 0 t/ha and 0.00013792 t/ha, respectively between the predicted and the actual values. The RFR and XGBR model predictions suggest that they might be susceptible to the high cardinality features. On the other hand, the DNN models, DNNR16 and DNNR64, maintained impressive residual error differences between the predicted and the actual. Further observations (see Figure 9 (b)) of the model's performance based on (2) are achieved by altering two explanatory variables, precipitation, and silt, individually. These variables were chosen because they exhibited a strong correlation, negatively and positively, to yield. As can be observed in Figure 4, silt is expected to have more impact on yield than precipitation, among other interacting variables. This implies that on the decrease of precipitation, the yield increases and vice versa. This implies that for every change in silt, the yield is expected to increase. The precipitation for Enugu and Plateau states changed from 133.5208333 to 13.5208333, and 99.125 to 9.125, respectively, while silt changed from 10.1666667 to 27.1666667, and 27.33333333 to 50.33333333, respectively. In Figure 9 (b) it can be observed that the decision-based models, RFR and XBGR, predictions practically remained unchanged for Enugu and Plateau states despite changes in precipitation and silt. This performance further reveals that when only a few explanatory variables are highly correlated to the outcome variable, the decision tree-based models might start becoming susceptible to high cardinality features. However, the DNN-based models utilized some (although possibly small) of the association between the explanatory and outcome variables to build a model that generalizes well to unforeseen changes. These are also in accordance with the expected impact of precipitation and silt on yield, as shown in Figure 4.

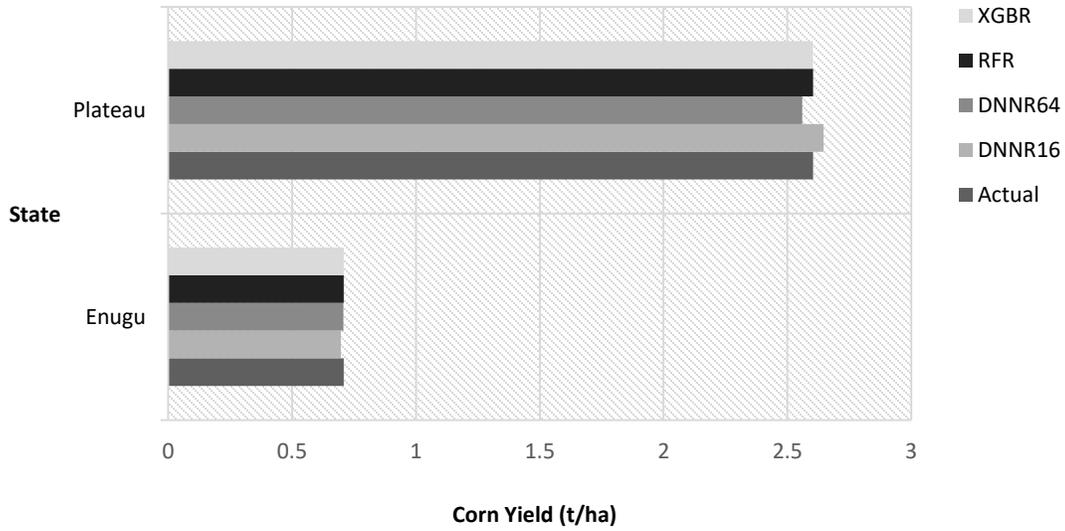

(a)

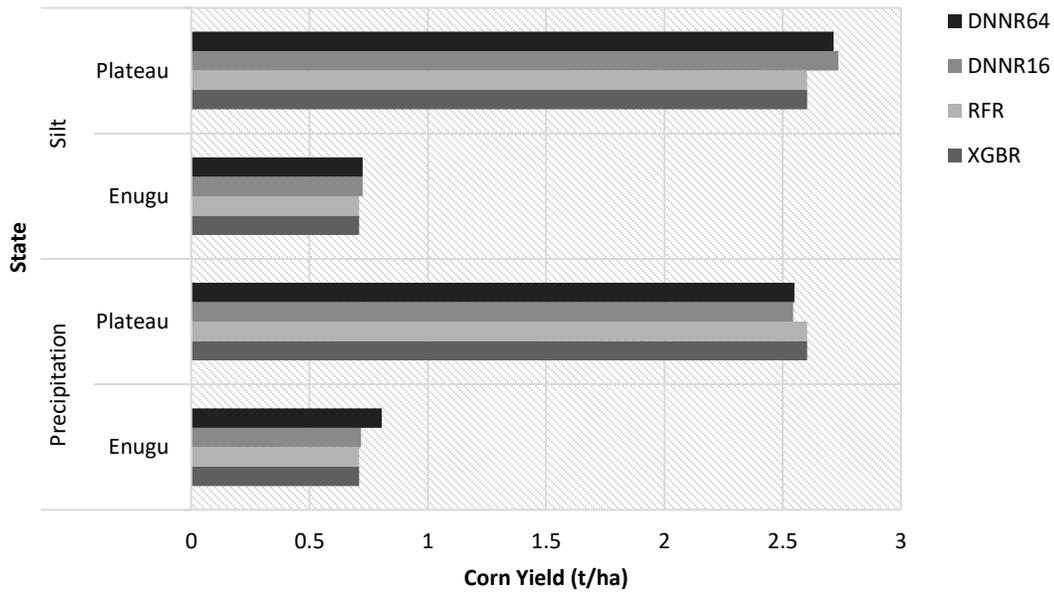

(b)

**Figure 9** A test of generalization of the predictive models to unseen and unforeseen data points. The models' prediction errors on the unseen data points are shown in (a) and unforeseen data points in (b). The unforeseen data points are created by increasing the values of two significant explanatory variables, precipitation, and silt, individually.

### 3.2.4 Significance of Feature Selection

Here, the aim is to observe the contribution of feature selection to the performance gains of the prediction models. The experiment carried out includes generating prediction errors with the DNN-based predictive models with and without feature selection. The outcome of the experiment can be visualized in Table 3, Figure 10 (a), and Figure 10 (b) and captures the performance of DNNR models on the entire test set and on only two sample points. It is interesting to observe that feature selection significantly contributed to the performance gains of the predictive models in both evaluation instances shown in Figure 10 (a) and 10 (b).

Table 3 Feature selection performance evaluation on the test set

| Mode | Model | RMSE (t/ha) | MAE (t/ha) | ARSE (t/ha) |
|------|-------|-------------|------------|-------------|
| w/o-FS | DNNR16 | 0.0296 | 0.01561 | 0.0139 |
| w-FS | DNNR16 | 0.0172 | 0.01195 | 0.00525 |
| w/o-FS | DNNR64 | 0.0296 | 0.01764 | 0.0119 |
| w-FS | DNNR64 | 0.0243 | 0.01745 | 0.00685 |

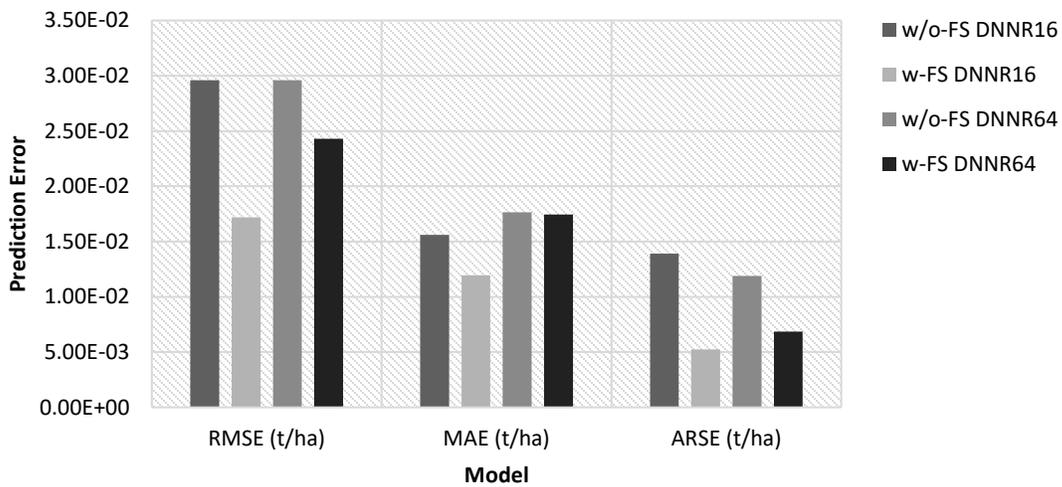

(a)

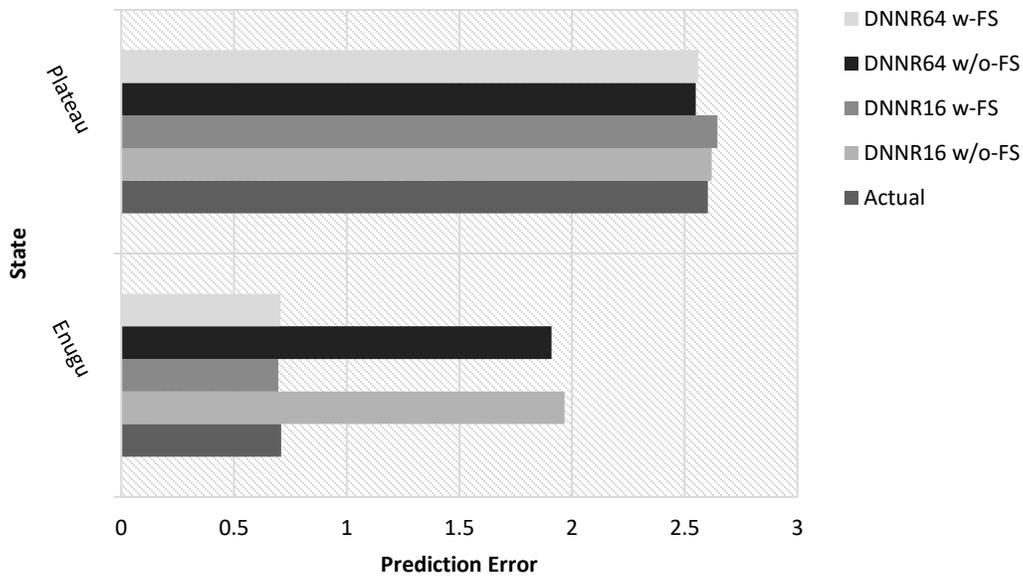

(b)

**Figure 10** Importance of feature selection in the prediction pipeline. Performance on the entire test set (a) and on two data points (b).

In Figure 10 (a), the ARSE, can be clearly seen to overcome the sensitivity and insensitivity of RMSE and MAE to outliers. A good example of insensitivity of MAE to outliers in the residual errors can be observed for the DNNR64 prediction errors recorded with and without feature selection. However, the difference in performance for both DNNR16 and DNNR64 became more obvious when the ARSE results are compared. The observed differences are 0.00865 t/ha and 0.00505 t/ha, respectively. As can be observed in Figure 10 (a), the significance of feature selection in the ML pipeline cannot be overemphasized.

Shown in Figure 10 (b) are performance results that provide insight into the significance of feature selection on close observation of two states, Enugu, and Plateau geo-positioned at opposing locations in Nigeria. The results with and without feature selection show that the impact of feature selection varies with the datapoint. For the Plateau state, feature selection did not appear to impact the prediction errors, with and without feature selection. However, a huge difference of about 1.25 t/ha prediction errors is observed for Enugu with and without feature selection.

### 3.3    Smallholder Farmer Decision Support System

The mobile phone is a technology that is commonly used across the globe and, if properly utilized, can become a powerful and portable artificial intelligence (AI) tool for performing intelligent tasks. Smallholder farmers have access to mobile phones and study [36] has shown that farmers use them to reach out to potential buyers for selling farm produce, and possibly for searching answers to farm challenges. This paper argues that mobile phone technology can be advanced with AI to transform the farming experience of smallholder farmers. A *proof-of-concept* mobile application is developed and implemented. It is termed "IntelliFarm" and comprises the following key functionalities: 1) farmer-market communication module to enable farmers to have direct access to buyers and vice-versa, and 2) education module subject to update on current best farming practices in agriculture, and 3) crop yield predictive model to help smallholder farmers make smart farming decisions for a given geo-location ahead of a planting season. The architecture of "IntelliFarm" is schematically shown in Figure 11 along with screenshots of the mobile application functionalities pertaining to yield prediction.

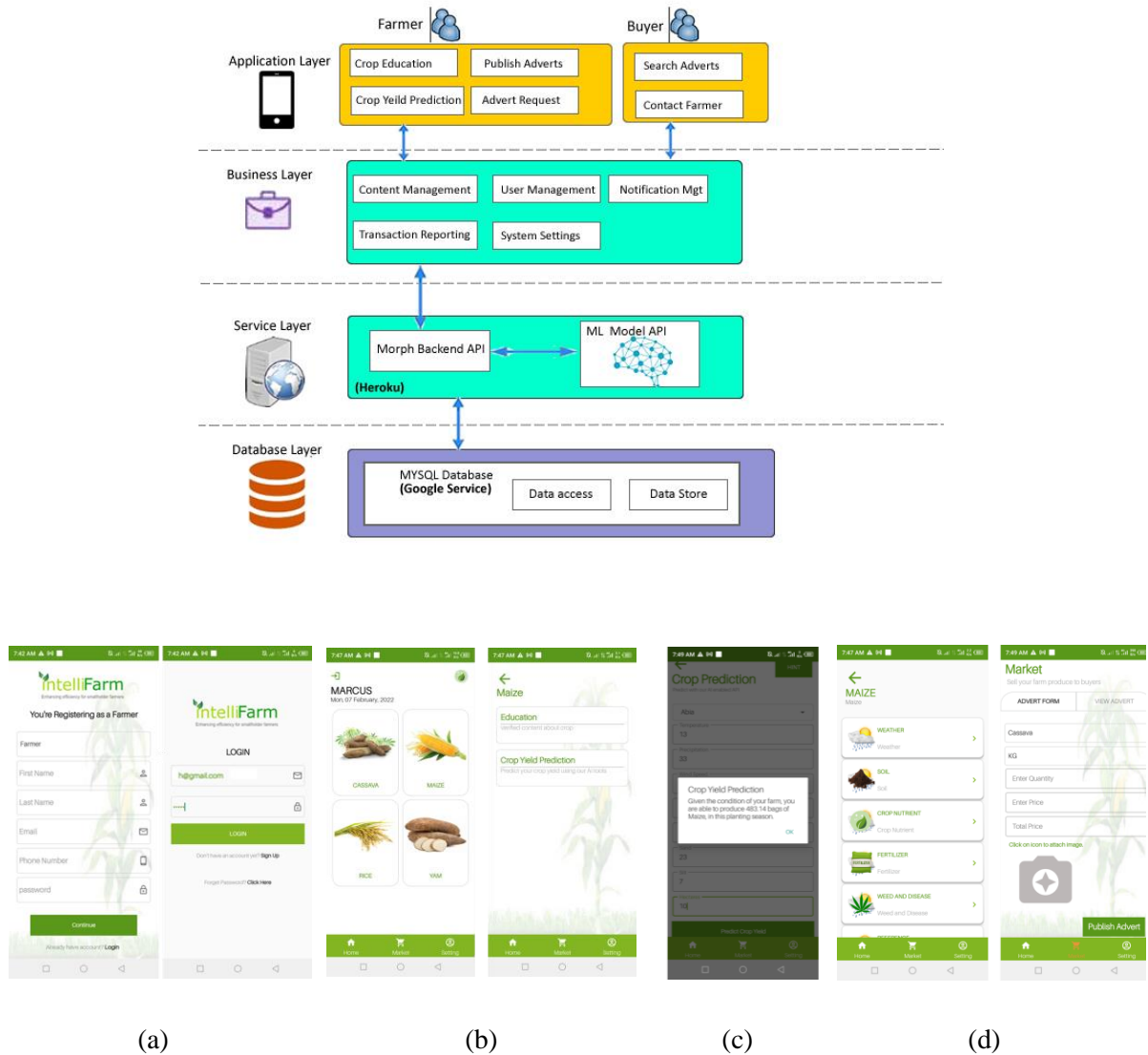

    (a)                 (b)                 (c)                 (d)

**Figure 11** The Decision support system mobile application architecture (upper) with four layers - the application, business, service, and database. The graphical user interface of the mobile application depicts the core functionalities (bottom). The farmer registration and login interface (a), welcome page (b), and prediction module (c), and education and market access (d).

The application architecture comprises the application layer, the business layer, the service layer, and the database layer. The application layer represents the users of the application, the farmer, and the buyer who can sign up and log in to access the application (see Figure 11 (a)) and perform core processes. These include sourcing educative materials for improving soil profile, creating farm produce profiles to enable the farmer to market produce to buyers, or predicting crop yield as shown in Figure 11 (b). The business layer is the backend of the application and ensures that the user accounts, notifications, and application usage reports are generated and stored. The service layer includes the application backend currently hosted under the Heroku cloud for hosting the corn yield predictive model. It is expected that when the smallholder farmer inputs the geolocation of their farm and the numeric values of the explanatory variables, they can estimate yield, *in bags per hectare*, for a given planting season. Prediction can be activated by clicking the prediction button. This sends a hypertext transfer protocol (HTTP) request through the application programmers' interface (API) to the Heroku Cloud server where the prediction model is hosted. Once the model receives the request, it runs an inference with the data and returns the model's prediction for corn yield, to the farmer. It should be noted that the deployed model in the early prototype is RFR, but any of the models can be used when the system is ready to be rolled out. The database layer is managed by the database management system, MySQL, hosted by Google Services. The data stored in the database are the farmer data, educative content, and images of farm produce for marketing to potential buyers. To help the smallholder farmer generate the required explanatory variables with ease, a third-party weather API can be utilized to retrieve temperature, precipitation, and wind speed data in real time, though this option is not explored in the current development. The education module is expected to provide tips on how local materials can be sourced for measuring soil profiles. Another valuable resource provided is a user manual, designed to help the smallholder farmer navigate the application with ease.

## 3.4    Discussion

The agricultural highlight of this paper is corn yield prediction that directly impacts the smallholder farmer. With the proposed corn yield prediction model, the smallholder farmer can be empowered to make informed decisions ahead of a planting season given weather, soil, and cultivation area. More importantly on a computational level the designed DNNR16 and DNNR64 models with outstanding capability in corn yield prediction and versatility under several scenarios of performance: overall prediction error, metrics evaluation, and generalization to unseen and unforeseen data. Although the RFR and XGBR models achieved the lowest prediction errors, it is found that they are reliant on cardinality features. This behavior became obvious through tests for generalization to unforeseen data, and since it is contrary to their performance on tabular data in the literature, we can only conclude that 1) there are small differences in the numeric values of the explanatory variables across different, 2) there is a non-linear relationship between the explanatory variables and the outcome variables, 3) there are too fee strongly correlated (negative and positive) explanatory variables to the outcome variables. Point (1) might have forced the decision tree-based models to depend on high cardinality features for variance across data points. Although the RFR and XGBR models achieved the smallest prediction errors with RMSE and MAE metrics, they could not accommodate any sort of changes to two weather and soil variables, precipitation, and silt - the variables among others identified via feature selection to show strong interaction with yield. Precipitation has a strong positive correlation, and silt has a strong negative correlation. On the contrary, the DNNR16 and DNNR64 models show superiority over the decision-based models. As is expected, a decrease in precipitation increases silt and an increase in precipitation reduces silt, and vice-versa. However, the degree of decrease or increase is not quantified in this paper because changes in the values could potentially assume any distribution due to the effect of climate change on environmental data. Additionally, the ARSE regression metric show to a good balance of RMSE and MAE. The ARSE has the potential to serve as a metric or loss function for regression tasks.

While there is no direct basis for comparison of this study to existing works [11]-[17] because different geolocations impose different impacts on crop farming and particularly the fact that the aim of the models differ, the similarity between them is worth noting. It is a known fact that neural networks are not great at quantifying the contributions of explanatory variables to outcome variables. However, this study took a different approach to identifying the features of most importance through the feature selection method and using that subset as the representative features for training the neural network. Therefore, these features will serve as the basis of comparison to existing literature as it pertains to the interaction of weather and soil variables with yield.

- The study in [14] identified the minimum temperature as the variable that is highly correlated to yield than maximum and average temperatures. This finding corresponds to the observation of this study. As shown in Figure 4, the Kendall correlation coefficient indicates that the minimum temperature is likewise highly correlated to yield than the maximum and average temperature, though it ranks as a weakly correlated feature among other correlated features such as cultivation area, silt, average precipitation, sand, soil pH, and wind speed. Note that these variables are arranged in their order of importance.

- The significance of precipitation to yield is heightened in this study which is similarly established in [14] and [12] to be significantly linked to yield.

- The soil variables on the other hand showed great variability across the literature. However, it is interesting to observe that silt and sand components of soil profiles, which are found in this study to be correlated to yield, coincide with the empirical results on the impact of soil analysis in agronomy exemplified by [37]. The study shows that sandy loam soil, a soil type formed by the combination of silt and sand soil, presents the best soil for the high-yield of corn plants.

Contrary to existing literature, this study provides a comprehensive decision support system to empower smallholder farmers to farm smartly. With a farmer's farm-to-market communication, third-party or middleman involvement in sales of farm produce is eliminated through the direct access to market module the decision support system provides. The educational resources provided in the mobile application are to help smallholder farmers to be well-informed to tackle farming challenges related to corn yield, fertilization, irrigation, and other maintenance practices. For instance, let's assume that the smallholder farmer after yield prediction decides that a predicted yield is below his/her expectation for a given planting season. They can explore the educational resources to gain insight into possible controllable factors such as soil pH, silt, and sand content of soil, to ensure that the expected yield can be achieved when interacting with weather factors that the smallholder farmer is unable to control.

The underlying strengths of this paper have so far been discussed however, there are several improvements to the methodology that can necessitate future research. This study is limited by the weather and soil variables the smallholder can easily acquire using available local resources for soil data, or third-party APIs for weather data. However, other variables such as organic radiation, water vapor, or saturated volumetric water content, might be relevant to yield and might be easily resourced by large-scale farms and can therefore expand the possibilities of the proposed model. Also, only corn yield is considered in this study, whereas there are other staple foods such as yam, cassava, guinea corn, and rice, with varying requirements for weather and soil interactions which can give way to further modelling.

## 4    Conclusion

This paper proposed a deep neural network regressor (DNNR) model where the depth and number of neurons of the hidden layer, amongst other hyperparameters, are structured to enable the network to model and learn the non-linear complex interaction between soil and weather data accurately. A new metric, ARSE, which combined the strengths of RMSE and MAE, was proposed for the regression task, and it forms a balance between RMSE, sensitive to outliers, and MAE, insensitive to outliers. With ARSE, the RFR, XGBR, DNNR16, and DNNR64 achieved yield prediction errors of 0.0001 t/ha, 0.001 t/ha, 0.0172 t/ha, and 0.0243 t/ha, respectively. However, when generalizability to unforeseen data due to changes to values of some of the explanatory variables was carried out, the DNNR models exhibited better modelling of the non-linear complexities of the environmental variables relative to their real-world agricultural expectations. The RFR and XGBR behaviors in such scenarios demonstrated their susceptibility to high cardinality problems. Further analysis revealed that a strong interaction existed between weather and soil variables, particularly with precipitation and silt variables shown to be strongly negatively and strongly positively correlated with yield. When precipitation value was reduced and silt value increased, yield increased and vice-versa. However, the degree of decrease or increase is not quantified in this study. Another highlight of the paper was the perspective for which the proposed yield predictive models are designed. They are fashioned to help the smallholder farmer make effective farming decisions that might have a wider and much more direct impact on alleviating food crises, in addition to the global purpose of monitoring food security and creating agricultural policies. Further advances are the design of a decision support system in the form of a mobile application. It integrated the proposed model and included educative and farmer-to-market access modules, to enable the smallholder farmer to farm smartly and intelligently. Future work centers around consolidating models for yield prediction of Africa's staple food.

### Declarations

- Funding

This research is funded by the UKRI/BBSRC Global Challenges Research Fund (GCRF) under the UK-Africa GCRF Agri-tech Catalyst Seeding Award with grant number 6429271 for a project titled "Artificial Intelligence (AI) Enhanced Smallholder Farm Software Tool."

The authors declare that there is no conflict of interest.

- Ethics approval

Not applicable.

- Consent to Participate

Not applicable.

- Consent for publication

Not applicable

- Data Availability

The datasets analysed during the current study are available at the following link: https://github.com/chollette/Corn-Yield-Prediction-Model-and-Mobile-Decision-Suport-SystemGithub.

- Code Availability

    The source code is available on GitHub in the following link: https://github.com/chollette/Corn-Yield-Prediction-Model-and-Mobile-Decision-Suport-SystemGithub.

- Authors Contributions

Conceptualization: C.C.O, L.S., M.S.; Methods, coding, model development, model deployment, writing – original draft: C.C.O; Literature review, C.C.O., M.O.L, L.S., M.S.; Data collection: C.C.O., M.O.L.; writing – review & editing: L.S., M.S., M.O.L., O.O: Business model design, and mobile application development and implementation: O.O., C.C.O., L.S., M.S.; Validation, evaluation, and funding acquisition: C.C.O., L.S., M.S.; Supervision, budgeting, project management: L.S., M.S; project administration: M.S.